\documentclass[letterpaper]{article}
\usepackage[final,dblblindworkshop]{neurips_2026}
\makeatletter
\renewcommand{\@notice}{}
\makeatother
\usepackage[T1]{fontenc}
\usepackage{orcidlink}
\hypersetup{hidelinks}
\usepackage[utf8]{inputenc}
\usepackage[T1]{fontenc}
\usepackage{hyperref}
\usepackage{url}
\usepackage{graphicx}
\usepackage{booktabs}
\usepackage{amsmath}
\usepackage{amssymb}
\usepackage{amsfonts}
\usepackage{nicefrac}
\usepackage{microtype}
\usepackage{xcolor}
\usepackage[T1]{fontenc}
\usepackage{orcidlink}
\hypersetup{hidelinks}

\usepackage{graphicx}
\usepackage{booktabs}
\usepackage{multirow}
\usepackage{array}
\usepackage{amsmath}
\usepackage{amssymb}
\usepackage[table]{xcolor}
\usepackage{colortbl}
\usepackage{enumitem}
\usepackage{algorithm}
\usepackage{algpseudocode}
\usepackage{amssymb}
\usepackage{algorithm}
\usepackage{algpseudocode}
\usepackage{xcolor}
\usepackage[most]{tcolorbox}

\definecolor{StageOneBorder}{HTML}{2E54A1}
\definecolor{StageOneTitle}{HTML}{B1D2E1}
\definecolor{StageOneBack}{HTML}{F7F9FD}
\definecolor{StageTwoBorder}{HTML}{588E31}
\definecolor{StageTwoTitle}{HTML}{ACD78E}
\definecolor{StageTwoBack}{HTML}{F7FBF4}
\definecolor{StageThreeBorder}{HTML}{C00000}
\definecolor{StageThreeTitle}{HTML}{EF949E}
\definecolor{StageThreeBack}{HTML}{FFF8F3}
\definecolor{FeedbackBorder}{HTML}{EE822F}
\definecolor{FeedbackTitle}{HTML}{F6D3AE}
\definecolor{FeedbackBack}{HTML}{FFF9F1}
\definecolor{NeutralBorder}{HTML}{2E54A1}
\definecolor{NeutralTitle}{HTML}{DAE3F5}
\definecolor{NeutralBack}{HTML}{FAFBFE}

\tcbset{
  appendixbox/.style={
    enhanced,
    breakable,
    boxrule=0.8pt,
    arc=1.0mm,
    outer arc=1.0mm,
    left=1.8mm,
    right=1.8mm,
    top=0.9mm,
    bottom=1.2mm,
    before skip=5pt,
    after skip=7pt,
    fonttitle=\bfseries\small,
    coltitle=black,
    title filled,
    colbacktitle=white,
    toptitle=0.8mm,
    bottomtitle=0.8mm,
    lefttitle=1.2mm,
    righttitle=1.2mm
  }
}

\tcbset{
  StageOne/.style={colframe=StageOneBorder,colback=StageOneBack,colbacktitle=StageOneTitle},
  StageTwo/.style={colframe=StageTwoBorder,colback=StageTwoBack,colbacktitle=StageTwoTitle},
  StageThree/.style={colframe=StageThreeBorder,colback=StageThreeBack,colbacktitle=StageThreeTitle},
  PoseStage/.style={colframe=StageTwoBorder,colback=StageTwoBack,colbacktitle=StageTwoTitle},
  FeedbackStage/.style={colframe=FeedbackBorder,colback=FeedbackBack,colbacktitle=FeedbackTitle},
  NeutralStage/.style={colframe=NeutralBorder,colback=NeutralBack,colbacktitle=NeutralTitle}
}

\newcommand{\PromptFileBox}[3]{%
\begin{tcolorbox}[appendixbox,#1,title={#2}]
\VerbatimInput[fontsize=\scriptsize,breaklines=true,breakanywhere=true]{#3}
\end{tcolorbox}%
}

\newcommand{\OutputFileBox}[2]{%
\begin{tcolorbox}[appendixbox,StageTwo,title={#1}]
\VerbatimInput[fontsize=\small,breaklines=true,breakanywhere=true]{#2}
\end{tcolorbox}%
}

\newcommand{\CodeFileBox}[2]{%
\begin{tcolorbox}[appendixbox,NeutralStage,title={#1}]
\VerbatimInput[fontsize=\small,breaklines=true,breakanywhere=true]{#2}
\end{tcolorbox}%
}

\newcommand{\ProtocolBox}[3]{%
\begin{tcolorbox}[appendixbox,#1,title={#2}]
#3
\end{tcolorbox}%
}

\usepackage{amsmath}
\usepackage{amssymb}
\usepackage{booktabs}
\usepackage{multirow}
\usepackage{array}
\usepackage{tabularx}
\usepackage{ragged2e}
\usepackage{fvextra}
\usepackage{float}
\usepackage{booktabs}
\usepackage{tabularx}
\usepackage{array}

\graphicspath{{latex/}}

\newcolumntype{L}[1]{>{\RaggedRight\arraybackslash}p{#1}}
\newcolumntype{Y}{>{\RaggedRight\arraybackslash}X}

\title{AssemState: Manual and Physical-State-Guided Reasoning for Zero-shot Furniture Assembly}
\author{
Zhiyuan Qi$^{1,3,*}$ \quad
Jierui Li$^{2,3}$\thanks{Equal contribution.} \quad
Yifan Shen$^{3}$ \quad
Cheng Qian$^{3}$ \quad
Jiateng Liu$^{3}$\\
$^{1}$Tsinghua University \quad
$^{2}$Xidian University \quad
$^{3}$University of Illinois Urbana-Champaign \\
\texttt{qzy24@mails.tsinghua.edu.cn}
\quad
\texttt{jerryli@stu.xidian.edu.cn}
}

\begin{document}
\maketitle

\begin{abstract}

Multimodal large language models (MLLMs) have made significant progress in visual understanding, but precise 3D spatial reasoning grounded in physical environments remains challenging. Furniture assembly requires recovering part and subassembly relations from diagrammatic manuals and translating semantic attachment relations into physically plausible 6D poses. To study this problem, we propose \textbf{AssemState}, a zero-shot framework for manual- and physical-state-guided furniture assembly, evaluated through two complementary subtasks. For assembly-structure recovery, it uses labeled parts, manual pages, and precomputed numbered step masks to identify boundary anchors, track subassemblies, and recover an assembly tree. For pose reasoning, given a correct atomic operation and preceding assembly state, it iteratively refines 6D pose estimates using rendered after-state feedback and constrained SE(3) corrections, with physical plausibility assessed through simulation-based release tests. Under the step-mask-conditioned protocol, AssemState achieves 62.80\% F1 and 53.92\% Tree Exact Match; removing anchor reasoning reduces these scores to 56.20\% and 45.10\%, respectively. On 243 independently evaluated part-level operations, refinement improves model-judge acceptance from 0 to 5.3\% and reduces mean Chamfer distance from 5.4111 to 1.7744. However, visually plausible poses can still exhibit collision, floating, mirror-orientation errors, incomplete seating, and wrong-side attachment. These results show improvements in structure recovery and selected local pose metrics, while contact-accurate furniture assembly remains challenging. The project page is available at \url{https://togekiss0603.github.io/AssemState}.
 
\end{abstract}
\begin{figure}[t]
    \centering
    \includegraphics[width=0.75\columnwidth]{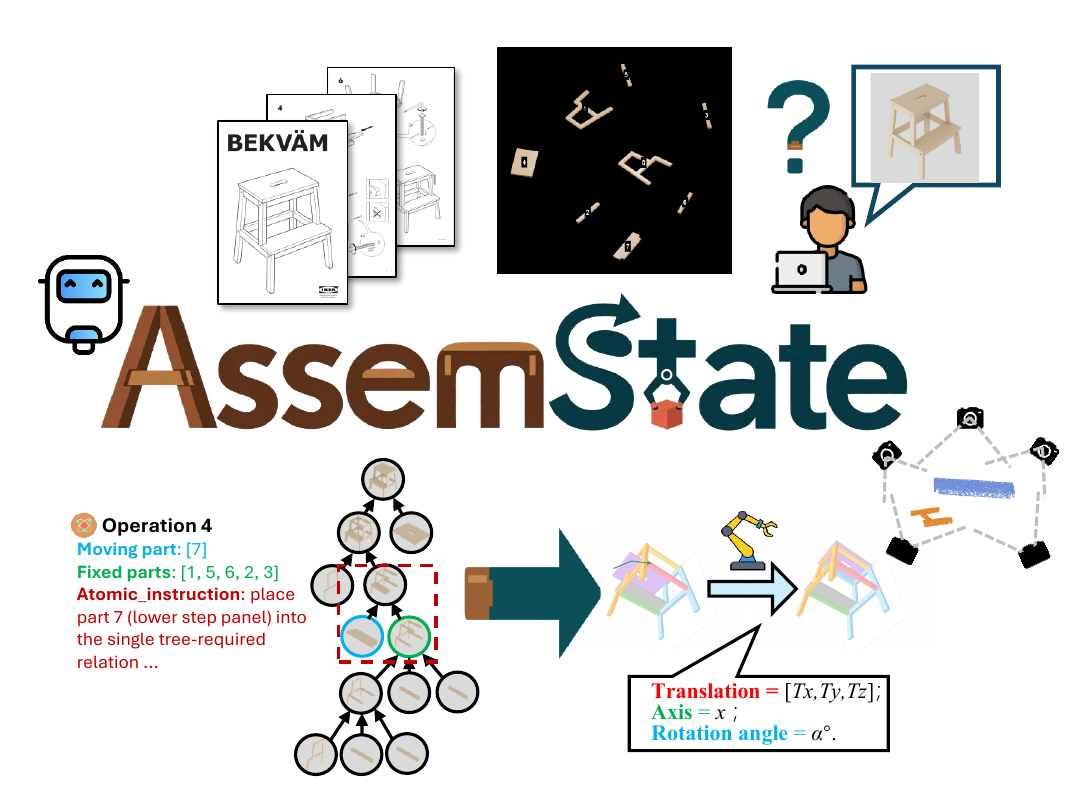}
    \caption{\textbf{Overview of AssemState.} The framework parses assembly manuals into structured part-level operations and an assembly tree, then leverages the evolving physical assembly state and after-state feedback to predict and refine 6D pose updates.}
    \label{fig:overall}
\end{figure}
\section{Introduction}

Diagrammatic furniture-assembly manuals specify parts, order, and attachment
relations, but converting them into physically meaningful actions remains
challenging. A manual page may involve multiple parts or interchangeable
components, while a semantic instruction such as attaching a frame to a panel
does not uniquely determine a contact-accurate 6D pose. A visually plausible
placement may still be geometrically invalid, unsupported, or attached to the
wrong side.

Recent datasets and environments, including 
\cite{Wang2022IKEAManual},  \cite{Zhang2023IAW}, 
\cite{Liu2024IKEAManualsAtWork}, and 
\cite{Heo2023FurnitureBench}, support research on manual understanding and
assembly. \cite{Tie2025Manual2Skill} and 
\cite{Tie2025Manual2SkillPP} study hierarchical and connector-aware assembly,
while closed-loop visual pose refinement has been explored for language-guided
6D rearrangement \cite{baik2026textguided}. These works motivate a controlled
study of two MLLM capabilities: recovering assembly structure from manuals and
reasoning about a current part-to-subassembly pose.

We present \textbf{AssemState}, a zero-shot framework with two separately
evaluated subtasks. For assembly-structure recovery, the model receives labeled
parts, manual pages, and precomputed numbered step masks, and uses boundary
anchors and subassembly tracking to recover a hierarchy. Mask generation is
outside the evaluation scope. For pose reasoning, AssemState receives a correct
current atomic operation and preceding assembly state, predicts a 6D placement,
renders the resulting after-state, diagnoses visible contact/alignment errors,
and applies constrained $SE(3)$ corrections. This protocol isolates local pose
reasoning from upstream tree errors and is not an end-to-end predicted-tree
rollout. We study two related reasoning problems while keeping their evaluation boundaries explicit.

Our contributions are:
\begin{itemize}
    \item We formulate manual-guided furniture assembly as two controlled
    zero-shot subtasks: step-mask-conditioned structure recovery and
    correct-operation-conditioned 6D pose reasoning.
    \item We introduce anchor-guided part/subassembly reasoning and iterative
    rendered after-state feedback for state-conditioned pose refinement.
    \item We show strong assembly-tree gains and improvements in selected local
    pose diagnostics, while documenting that strict contact-accurate pose
    recovery remains unsolved.
\end{itemize}
\section{Related Work}

\paragraph{MLLMs for procedural spatial reasoning.}
Prior research demonstrates the promising abilities of MLLMs and VLA models in interpreting and reasoning the spatial-temporal meaning of motion actions continuously. ~\cite{driess2023palme,zitkovich2023rt2,kim2024openvla} connect multimodal observations with robot actions, enabling generalist language-conditioned manipulation across diverse tasks.  Beyond action generation, ~\cite{chen2024spatialvlm,huang2023voxposer} improve spatial grounding by injecting metric spatial supervision or composing language-conditioned 3D value maps for manipulation. For assembly-specific reasoning, ~\cite{jing2026assemlm} integrate standard manuals, textual instructions and point clouds to predict static poses, while ~\cite{pun2025brickgpt,wen2026bricksim} emphasize the importance of buildability and physical stability with units of any shape.

\paragraph{6D pose estimation.}
Accurate 6D pose estimation is the geometric foundation of assembly, since even a locally plausible operation can fail if the target pose is misaligned. ~\cite{wen2024foundationpose} unify novel-object pose estimation and tracking under both model-based and model-free settings, improving generalization to unseen objects through synthetic training and pose refinement. ~\cite{liu2026activepose} further addresses viewpoint-induced ambiguity by combining pose-hypothesis uncertainty with MLLM-guided direction selection, showing the value of active perception for reliable pose recovery. ~\cite{baik2026textguided} iteratively refine object poses through rendered feedback. ~\cite{zhang2026vlm2pose} ground language-specified viewpoints into explicit camera poses for multi-view spatial reasoning. ~\cite{chi2023diffusionpolicy} complement these methods by modeling precise visuomotor control as action diffusion. Despite these advances, existing methods mainly address pose estimation, spatial reasoning, action generation, or physical feasibility in isolation, leaving long-horizon assembly from ambiguous manuals underexplored.

\begin{figure*}[t]
    \centering
    \includegraphics[width=0.85\textwidth]{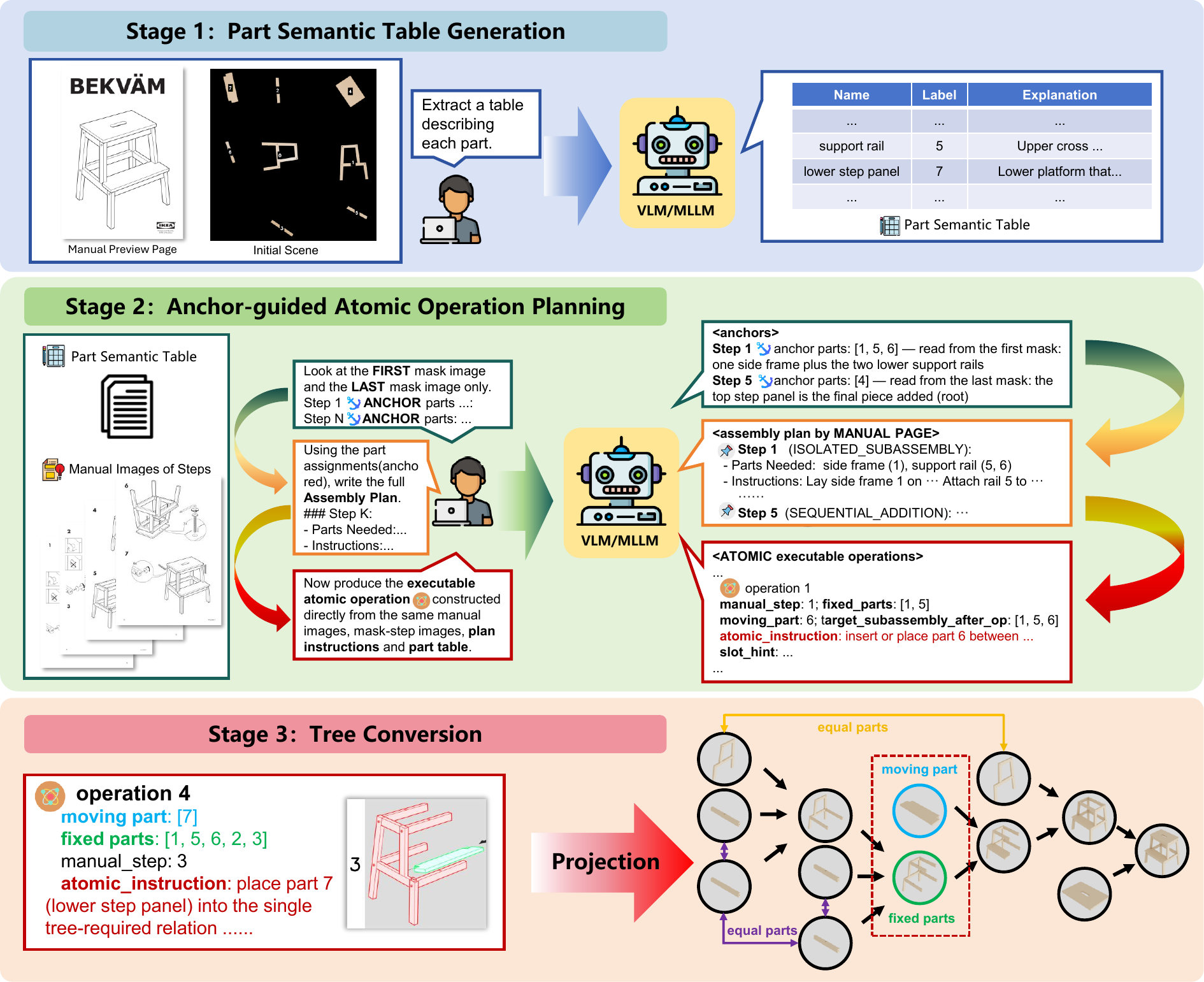}
\caption{\textbf{Anchor-guided manual-to-structure reasoning.}
The manual-side pipeline uses part semantics and ordered highlighted step masks to recover step-level part/subassembly structure and an assembly tree. Atomic moving-part operations are a downstream interface obtained from the hierarchy rather than being assumed to correspond one-to-one with manual pages.}
    \label{fig:manual_to_operation}
\end{figure*}

\section{Problem Setting}

\paragraph{Manual-side assembly-structure recovery.}
Let $\mathcal{P}_0=\{p_0,\ldots,p_{N-1}\}$ denote $N$ labeled furniture parts in the initial scene, and let $\mathcal{I}=\{I_0,\ldots,I_{M-1}\}$ denote the ordered manual pages. The current implementation additionally receives ordered, precomputed numbered step masks $\mathcal{H}=\{H_0,\ldots,H_{M-1}\}$, where the highlighted parts provide the step-localized part cues used by the anchor planner. We treat $\mathcal{H}$ as an observed input to the manual-side benchmark; generating these masks from raw manuals is outside the scope of the reported evaluation.

The goal is to recover a hierarchical assembly tree $\mathcal{G}$ whose leaves are part IDs and whose internal nodes represent intermediate subassemblies. A manual step may introduce several parts or form an isolated subassembly, so we do not assume a one-to-one mapping between the $M$ manual pages and $N-1$ atomic moving-part operations. Instead, the manual pipeline first recovers step-level part/subassembly organization and then converts that organization to the hierarchy $\mathcal{G}$.

\paragraph{Atomic-operation interface.}
For downstream pose reasoning, an atomic current operation is represented as
\[
 a_q=\bigl(\mathcal{P}_{q,\mathrm{moving}},\mathcal{P}_{q,\mathrm{fix}},r_q,h_q\bigr),
\]
where $\mathcal{P}_{q,\mathrm{moving}}$ is the moving part, $\mathcal{P}_{q,\mathrm{fix}}$ is the current fixed subassembly, $r_q$ is a semantic attachment relation, and $h_q$ is a pose hint. Conceptually, such operations can be instantiated by traversing an assembly hierarchy. In the reported pose benchmark, however, $a_q$ and the preceding assembly state are conditioned as correct in order to isolate pose reasoning from upstream tree errors.

\paragraph{Correct-operation-conditioned pose reasoning.}
Given the correct current operation $a_q$ and preceding state $\mathcal{S}_q$, the pose subtask predicts a 6D pose
\[
\hat T_q=(\hat R_q,\hat t_q)\in SE(3)
\]
for the moving part. 


\section{Methods}

AssemState contains \textbf{anchor-guided manual-to-structure reasoning} and \textbf{rendered after-state pose refinement}. The supplementary material gives the exact schedule and preserved prompt snapshots.

\subsection{Anchor-Guided Manual-to-Structure Reasoning}
As shown in Fig.~\ref{fig:manual_to_operation}, the manual-side pipeline consists of three conceptual stages implemented by a short sequence of MLLM calls and a tree-conversion step.

\subsubsection*{Stage 1: Part Semantic Table Generation}
The first call predicts number--name pairs from a top-down numbered scene, available annotated scene views, and a manual preview page. A second call receives this part table together with the scene/views and the full manual-page sequence, and augments each part with a brief functional description. The result is a semantic table
\[
B=\{(\mathrm{id}_i,\mathrm{name}_i,\mathrm{desc}_i)\}_{i=0}^{N-1}.
\]
This stage grounds the model to the labeled structural parts observed in the scene and reduces the tendency to add manual-only hardware or tools as furniture parts.

\subsubsection*{Stage 2: Anchor-Guided Assembly-Structure Planning}
A single manual page can involve multiple parts, so this stage does not directly equate pages with atomic moving-part operations. Instead, the anchor planner receives the semantic table $B$, the numbered scene, and all ordered numbered step masks $\mathcal{H}$. Within one MLLM request, it performs three pieces of reasoning: (i) identify the first- and last-step boundary anchors, (ii) assign highlighted part IDs to steps while tracking carried-over or isolated subassemblies, and (iii) produce a textual assembly plan.

Let $L=\{l_0,\ldots,l_{M-1}\}$ denote the resulting step-level plan. Abstractly,
\[
L=\Phi(V,\mathcal{H},B),
\]
where $V$ denotes the numbered scene. The highlighted step masks are an explicit input to $\Phi$ and provide strong part-to-step localization cues; AssemState's contribution at this stage is the anchor/subassembly reasoning conditioned on those cues, not upstream mask generation. The planner also checks that part IDs are not omitted or reused across its step assignments.

\subsubsection*{Stage 3: Assembly-Tree Conversion and Operation Interface}
The Stage-2 text is converted to a nested-list hierarchy $\mathcal{G}$. The implementation first attempts a rule-based parse and uses a text-only MLLM conversion prompt when the structured parse is unavailable. Leaves of $\mathcal{G}$ correspond to atomic parts, while internal nodes represent intermediate subassemblies and the outermost node represents the completed hierarchy.

Traversing $\mathcal{G}$ provides a natural way to instantiate an atomic operation tuple
\[
 a_q=\bigl(\mathcal{P}_{q,\mathrm{moving}},\mathcal{P}_{q,\mathrm{fix}},r_q,h_q\bigr)
\]
for pose reasoning. This tree-to-operation interface is conceptual in the overall framework. The reported pose benchmark deliberately conditions on the correct current operation/state rather than directly consuming the predicted \texttt{tree.json}; this separates spatial pose errors from upstream structure-recovery errors.

\begin{figure*}[t]
    \centering
    \includegraphics[width=0.99\textwidth]{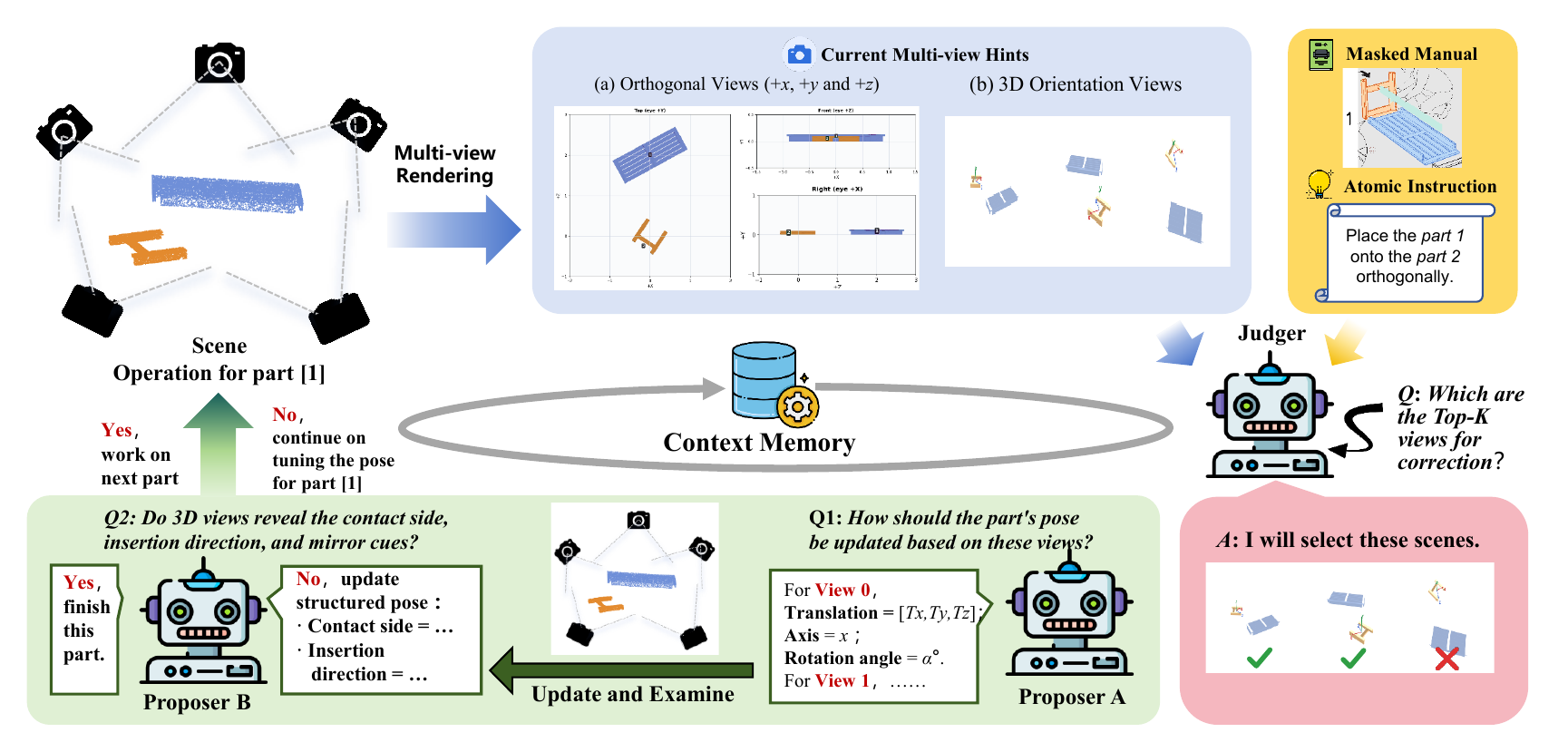}
    \caption{\textbf{Iterative after-state feedback refinement.} Multi-view renderings of the updated current-operation state guide successive state-conditioned corrections. The reported benchmark evaluates each operation independently under a correct-operation/preceding-state protocol.}
    \label{fig:pose_estimation_pipeline}
\end{figure*}

\subsection{Target-Free 6D Pose Reasoning and Refinement}
For a correct current operation $a_q$, the model predicts a candidate 6D pose
\[
\hat{T}_{q}^{0}=(\hat{R}_{q}^{0},\hat{t}_{q}^{0})\in SE(3)
\]
for the moving part, without access to a ground-truth target pose or target-state image.



\paragraph{Iterative after-state feedback.}
At refinement round $r$, the current estimate is applied and the resulting after-state is rendered from multiple views:
\[
\hat{\mathcal{S}}_q^{r}
=
\hat{T}_q^{r}[m_q]\cup(\mathcal{S}_q\setminus\{m_q\}),
\]
where $m_q$ denotes the moving part. A diagnosis of the rendered after-state records whether the current candidate is acceptable and, when rejected, identifies a failure type and residual correction cues such as wrong side, wrong mirror, floating, not fully seated, collision, or orientation error. 


Rather than generating a new pose independently from the original scene, the next proposal is conditioned on the current state and the previous diagnosis, predicting an incremental correction
\[
\delta T_q^{r}=(\delta R_q^{r},\delta t_q^{r})\in SE(3),
\qquad
\hat T_q^{r+1}=\delta T_q^{r}\hat T_q^{r}.
\]
The prompt constrains the correction according to the diagnosed error: for example, seating/floating/collision errors favor local translations, whereas wrong-side or mirror errors may require a larger side/orientation change. We use three refinement rounds in the main reported setting. The final prediction is $\hat T_q^{R}$ with $R=3$.

\section{Experiments}

We evaluate five aspects of AssemState: (1) manual-side assembly-structure recovery, (2) correct-operation-conditioned target-free 6D pose estimation, (3) simulation-based release diagnostics, (4) component ablations, and (5) failure analysis.

\paragraph{Evaluation scope.}
The two main evaluations are intentionally separated. The manual-side results are conditioned on the precomputed numbered step masks documented in the appendix and therefore measure structure reasoning under highlighted step-part cues rather than raw-manual mask generation. The pose results condition on a correct current operation and preceding assembly state, and therefore measure current-operation spatial reasoning rather than end-to-end propagation from a predicted tree. These boundaries are summarized in the appendix together with the exact prompt/state schedule.

\subsection{Manual-to-Structure Recovery}

\paragraph{Setup and metrics.}
Using GPT-5~\cite{openai2025gpt5systemcard}, we evaluate recovery of the reference assembly hierarchy. Precision, Recall, and F1 (\%) are computed from predicted versus reference parent--child relations, while Tree Exact Match (Tree EM, \%) requires the complete hierarchy to match under the evaluation protocol. The input protocol includes the labeled scene, manual context, and ordered numbered step masks; upstream mask construction is not part of this evaluation.

\begin{table}[t]
\centering
\small
\setlength{\tabcolsep}{5.5pt}
\begin{tabular*}{\linewidth}{@{\extracolsep{\fill}}lcccc@{}}
\toprule
\textbf{Method} & \textbf{Prec.} & \textbf{Rec.} & \textbf{F1} & \textbf{Tree EM} \\
\midrule
SingleStep        & 10.78 & 10.78 & 10.78 & 10.78 \\
GeoCluster        & 13.37 & 13.36 & 13.15 &  3.92 \\
Manual2Skill      & 37.45 & 36.41 & 36.83 & 27.45 \\
\midrule
\textbf{Ours} w/o Anchors
                  & \underline{57.39} & \underline{55.90} & \underline{56.20} & \underline{45.10} \\
\textbf{Ours}
                  & \textbf{62.34} & \textbf{62.46} & \textbf{62.80} & \textbf{53.92} \\
\bottomrule
\end{tabular*}
\caption{Assembly-structure recovery results (\%) under the reported manual-side evaluation protocol. Best results are in \textbf{bold} and second-best are \underline{underlined}.}
\label{tab:tree_generation}
\end{table}

\paragraph{Baselines.}
We include SingleStep, which predicts a flattened ordered tree; GeoCluster, which uses a trained DGCNN~\cite{phan2018dgcnn}; and Manual2Skill~\cite{Tie2025Manual2Skill}, an MLLM-based approach to furniture assembly.

\paragraph{Results.}
Under the reported step-mask-conditioned evaluation protocol, AssemState achieves an F1 score of 62.80\% and a Tree Exact Match (Tree EM) rate of 53.92\%, as shown in Table~\ref{tab:tree_generation}. Removing anchor reasoning reduces F1 to 56.20\% and Tree EM to 45.10\%, corresponding to decreases of 6.60 and 8.82 percentage points, respectively. These results demonstrate the contribution of boundary anchors and subassembly tracking to organizing highlighted step-part cues and recovering assembly hierarchies. We restrict these findings to structure recovery under the supplied step-mask-conditioned protocol; the experiment does not evaluate upstream mask generation.

\begin{table*}[t]
\centering
\small
\setlength{\tabcolsep}{3pt}
\renewcommand{\arraystretch}{1.10}

\makebox[\textwidth][c]{%
\textbf{(a) Correct-operation-conditioned target-free 6D pose estimation}}
\par\vspace{3pt}

\begin{tabular*}{\textwidth}{@{\extracolsep{\fill}}llcccc@{}}
\toprule
\textbf{Backbone} &
\textbf{Setting} &
\shortstack{\textbf{Judge Acc.}\\\textbf{(\%) $\uparrow$}} &
\shortstack{\textbf{RMSE}$_t$\\$\downarrow$} &
\shortstack{\textbf{GD}$_{\deg}$\\$\downarrow$} &
\shortstack{\textbf{CD / Med CD}\\$\downarrow$} \\
\midrule

\multirow{3}{*}{GPT-5}
& Text-only baseline
& 0.0
& 2.4322
& 123.02
& 15.7263 / 10.1914 \\

& Open6DOR-style baseline
& 0.0
& \textbf{1.8747}
& \textbf{120.84}
& 5.4111 / 0.9391 \\

& \textbf{Ours: Self-check 3-round}
& \textbf{5.3}
& 1.8898
& 121.52
& \textbf{1.7744 / 0.8112} \\

\midrule

\multirow{3}{*}{\shortstack[l]{Claude\\Haiku 4.5}}
& Text-only baseline
& 0.8
& 2.4315
& 125.61
& 18.7281 / 13.8936 \\

& Open6DOR-style baseline
& 4.5
& \textbf{1.8887}
& \textbf{123.20}
& 10.8826 / 3.2725 \\

& \textbf{Ours: Self-check 3-round}
& \textbf{11.5}
& 1.9760
& 123.69
& \textbf{10.6708 / 1.8770} \\

\bottomrule
\end{tabular*}

\vspace{7pt}

\makebox[\textwidth][c]{%
\textbf{(b) Simulation-based release validation}}
\par\vspace{3pt}

\begin{tabular*}{\textwidth}{@{\extracolsep{\fill}}llccc@{}}
\toprule
\textbf{Backbone} &
\textbf{Setting} &
\shortstack{\textbf{Release Stability}\\\textbf{(\%) $\uparrow$}} &
\shortstack{\textbf{Mean Drift}\\$\downarrow$} &
\shortstack{\textbf{Mean Z Drop}\\$\downarrow$} \\
\midrule

\multirow{3}{*}{GPT-5}
& Text-only baseline
& 16.5
& 0.4249
& 0.1696 \\

& Open6DOR-style baseline
& 37.0
& 0.2463
& 0.1058 \\

& \textbf{Ours: Self-check 3-round}
& \textbf{39.5}
& \textbf{0.1754}
& \textbf{0.0632} \\

\midrule

\multirow{3}{*}{\shortstack[l]{Claude\\Haiku 4.5}}
& Text-only baseline
& 12.3
& 0.4254
& 0.1859 \\

& Open6DOR-style baseline
& \textbf{33.7}
& 0.3266
& 0.1328 \\

& \textbf{Ours: Self-check 3-round}
& \textbf{33.7}
& \textbf{0.3109}
& \textbf{0.1262} \\

\bottomrule
\end{tabular*}

\caption{
\textbf{Current-operation pose reasoning and release validation on the correct-tree subset.}
All rates are computed over 243 independently evaluated atomic operations.
The pose benchmark conditions on a correct current operation and preceding state;
it is not a predicted-tree end-to-end rollout.
PA@0.01 is 0.0000 for every setting and is omitted from the table for compactness.
}
\label{tab:pose_sim_result}
\end{table*}

\subsection{Target-Free 6D Pose Estimation}

\paragraph{Setup and metrics.}
We evaluate the pose subtask on 55 correct-tree cases comprising 243 atomic current operations. Each operation is scored independently because an early incorrect placement would otherwise invalidate later states. After scoring, the moving part is reset to its ground-truth pose only to construct the context for the next operation; the target pose is never exposed to the MLLM as an input or target image. We report five metrics: Judge acceptance, pose accuracy at a strict tolerance\(PA @0.01\), translation RMSE$_t$, geodesic rotation distance (GD), and Chamfer distance (CD). Formal definitions and evaluation details are provided in the Appendix.

\paragraph{Baselines.}
The text-only baseline predicts from the atomic-operation description without current-state images. The Open6DOR-style \cite{qi2025sofar} baseline predicts a structured one-shot 6D update. AssemState adds multi-view state grounding and three rounds of after-state feedback with constrained $SE(3)$ correction. Aggregate results are reported under the backbone labels preserved with the experiment tables; the appendix separately documents the exact API identifier available for one representative historical trace.

\paragraph{Results.}
Table~\ref{tab:pose_sim_result}(a) shows a mixed but informative pattern. Relative to the Open6DOR-style baseline, three-round self-check increases model-judge acceptance from 0.0\% to 5.3\% for GPT-5 and from 4.5\% to 11.5\% for Claude Haiku 4.5 (\cite{anthropic2025claudehaiku45}). Mean CD also decreases from 5.4111 to 1.7744 for GPT-5, while Claude's median CD decreases from 3.2725 to 1.8770. In contrast, translation RMSE and geodesic rotation distance do not consistently improve, and PA@0.01 remains 0 for all methods. We therefore interpret the refinement as improving selected local/geometric diagnostics and model-judge contact consistency, not as solving target-free 6D assembly pose estimation.

\begin{figure*}[t]
    \centering
    \includegraphics[width=0.8\textwidth]{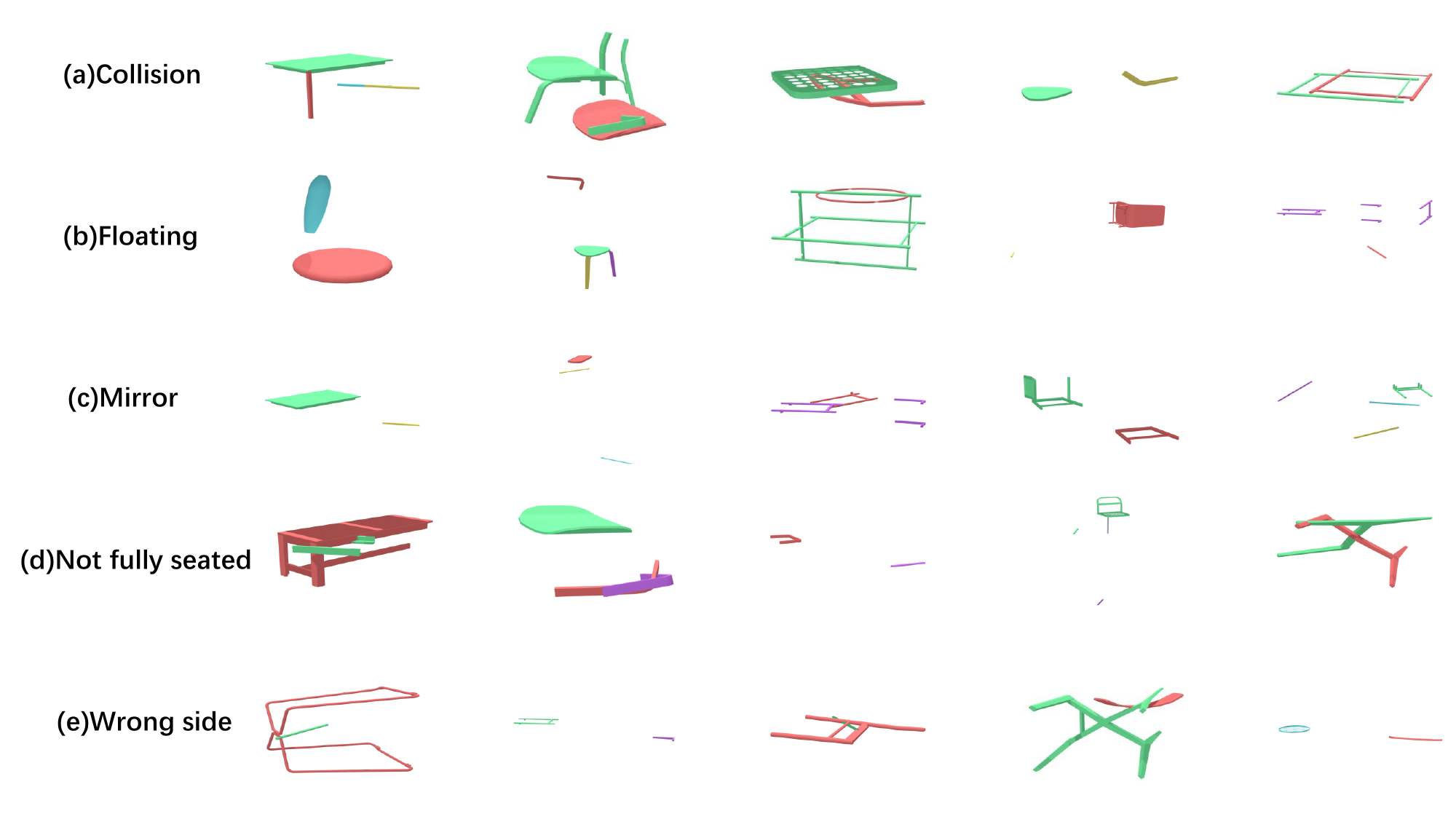}
    \caption{Representative failure cases of the final self-check setting, including collision, floating, mirror, not-fully-seated, and wrong-side errors.}
    \label{fig:failure_cases}
\end{figure*}

\subsection{Simulation-Based Validation}

We additionally release the selected current-operation candidate in simulation and measure whether it remains stable. These diagnostics complement, rather than replace, strict pose accuracy.

For GPT-5, Table~\ref{tab:pose_sim_result}(b) shows that self-check reduces mean drift from 0.2463 to 0.1754 relative to the Open6DOR-style baseline, while release stability increases modestly from 37.0\% to 39.5\%. For Claude Haiku 4.5, drift and vertical drop decrease slightly, but binary release stability remains unchanged at 33.7\%. Thus, after-state refinement can improve some physical plausibility diagnostics, but the effect is not uniform across metrics or backbones.

\subsection{Ablation Study}

\begin{table}[t]
\centering
\small
\setlength{\tabcolsep}{1mm}
\begin{tabular}{@{}p{0.37\columnwidth}ccc@{}}
\toprule
\textbf{Setting} &
\begin{tabular}[c]{@{}c@{}}\textbf{Judge}\textbf{ Accept.}$\uparrow$\end{tabular} &
\textbf{CD}$_\downarrow$ &
\begin{tabular}[c]{@{}c@{}}\textbf{Release}\\\textbf{Stability}$\uparrow$\end{tabular} \\
\midrule
Open6DOR-style
& 0/243
& 5.4111
& 90/243 = 37.0\% \\
+ Multi-view Structured Init.
& 1/243
& 5.3113
& \textbf{109/243 = 44.9\%} \\
+ Contact-aware Self-check
& \textbf{13/243}
& \textbf{1.7744}
& 96/243 = 39.5\% \\
\bottomrule
\end{tabular}
\caption{Cumulative pose ablation. Judge acceptance is a model-based diagnostic; each row adds one component to the previous setting.}
\label{tab:pose_ablation}
\end{table}

We use a cumulative ablation to separate coarse multi-view grounding from contact-aware after-state feedback. The three reported quantities are judge acceptance, CD, and release stability.
 Table~\ref{tab:pose_ablation} shows that multi-view initialization has its clearest effect on release stability (37.0\% to 44.9\%), whereas the subsequent contact-aware self-check has its clearest effect on judge acceptance (1/243 to 13/243) and CD (5.3113 to 1.7744). Release stability decreases from the multi-view-only setting to 39.5\%, so the ablation does not support a claim of monotonic improvement across all objectives. Instead, it suggests that the components affect different aspects of current-operation pose quality.

\subsection{Failure Analysis}

Figure~\ref{fig:failure_cases} shows representative failures of the final self-check setting. We use GPT-5 to categorize failed rendered after-states according to the manual instruction and expected contact relation; these labels are a model-assisted diagnostic rather than human ground-truth annotations. The failures follow a clear progression. The model usually first localizes a
plausible coarse region for the moving part, but this rough localization does not
by itself guarantee a physically valid assembly. The residual errors fall into the
following recurring categories:
\begin{itemize}
\item \textbf{Collision.} The predicted pose can look reasonable from some views
      yet intersects the fixed subassembly, showing that visual plausibility does
      not ensure non-penetration.
\item \textbf{Floating.} The moving part lacks stable support and drifts or drops
      after release, so a visually acceptable placement still violates the required
      support relation.
\item \textbf{Wrong-side.} The model identifies the correct local region but
      attaches the part to the wrong connector face, reflecting the difficulty of
      resolving contact topology once the coarse region is reached.
\item \textbf{Mirror.} The part is placed in a flipped (mirrored) orientation; such
      errors are especially frequent under symmetry or partial occlusion.
\item \textbf{Not-fully-seated.} Even when the correct side is selected, the part
      reaches the contact region but fails to complete the final insertion or tight
      seating required by the assembly.
\end{itemize}


\section{Conclusion}

We presented \textbf{AssemState} as a zero-shot study of two complementary but separately evaluated capabilities for manual-guided furniture assembly: step-mask-conditioned assembly-structure recovery and correct-operation-conditioned target-free 6D pose reasoning. Boundary anchors and subassembly tracking improve assembly-tree recovery under the manual-side input protocol, while rendered after-state feedback can improve model-judge acceptance and Chamfer distance for independently evaluated current operations. At the same time, failure cases continue to exhibit collision, floating, mirror, wrong-side, and incomplete-seating errors. These results support the use of MLLMs for structured manual reasoning and iterative local correction, while showing that reliable end-to-end, contact-accurate furniture assembly remains an open problem.


\bibliographystyle{plainnat}
\bibliography{latex/custom}

\clearpage
\appendix
\section{Appendix}

\section{Metrics and Physical Diagnostic}
\label{app:metrics}
\subsection{Pose Discrepancy}
Given reference and predicted poses in a common frame, $(R^*,t^*)$ and $(\hat R,\hat t)$, the translation-coordinate RMSE is
\[
\operatorname{RMSE}_{t,i}=\sqrt{\frac{1}{3}\|\hat t_i-t_i^*\|_2^2}.
\]
The reported mean is the mean of per-operation RMSE values, not a fresh root over all products' coordinates. Rotation error is
\[
\operatorname{GD}_i=\frac{180}{\pi}
\arccos\!\left(\operatorname{clip}\left[
\frac{\operatorname{tr}((R_i^*)^\top\hat R_i)-1}{2},-1,1\right]\right).
\]
Continuous shape symmetry can make orientation error differ from functional correctness. No universal continuous-symmetry correction is claimed for the table's raw GD.

The visual-agent metric script implements bidirectional squared nearest-neighbor Chamfer:
\[
\operatorname{CD}(X,Y)=
\frac{1}{|X|}\sum_{x\in X}\min_{y\in Y}\|x-y\|_2^2+
\frac{1}{|Y|}\sum_{y\in Y}\min_{x\in X}\|y-x\|_2^2.
\]
It uses corresponding OBJ vertex subsampling and a KD-tree query, rather than necessarily uniform sampling over surface area. Its current defaults are at most 2048 corresponding vertices and seed 0; alignment can be global, base-anchored, or none. These are source-level defaults, not a recovered command for every historical table row. Another legacy metric module in the same repository uses unsquared nearest-neighbor distances, so it is unsafe to treat every function labeled Chamfer as interchangeable.

\textbf{Pose Accuracy (PA$@$0.01)} is a strict, binary per-operation success criterion. An atomic
operation $i$ is counted as correct only if its part-level discrepancy falls
below a fixed tolerance $\tau = 0.01$:
\[
\operatorname{PA@}\tau=\frac{1}{N}\sum_{i=1}^{N}
\mathbf{1}\left[\operatorname{CD}_i < \tau\right],\qquad \tau = 0.01,
\]
where $\operatorname{CD}_i$ is the bidirectional squared Chamfer distance
defined above and $N$ is the number of evaluated atomic operations. In contrast
to the mean CD, RMSE$_t$, and GD, which are continuous discrepancy measures,
PA@0.01 reports the fraction of operations placed within the tight tolerance
required for contact-accurate assembly, and therefore serves as our strictest
pose-correctness metric.

\subsection{Judge Acceptance}
\label{app:judge_acceptance}

Historical rows use the stored Boolean \texttt{accepted}. The archived
judge outputs contain an acceptance decision together with a coarse failure
diagnosis and correction hints. A representative response schema is:

\begin{tcolorbox}[
    colback=blue!3,
    colframe=blue!45!black,
    boxrule=0.6pt,
    arc=2pt,
    left=6pt,
    right=6pt,
    top=5pt,
    bottom=5pt,
    title=\textbf{Archived judge-response schema},
    fonttitle=\small,
]
{\small\ttfamily
\{\\
\hspace*{1em}"Accept": "Yes or No",\\
\hspace*{1em}"Error type": "wrong\_side / wrong\_mirror / floating / collision / ...",\\
\hspace*{1em}"Residual direction": "textual correction direction",\\
\hspace*{1em}"Contact": "Correct / Incorrect / Unclear",\\
\hspace*{1em}"Orientation": "Correct / Incorrect / Unclear",\\
\hspace*{1em}"Reasoning": "visual evidence",\\
\hspace*{1em}"Next candidate hints": ["..."],\\
\hspace*{1em}"Confidence": 0.0\\
\}
}
\end{tcolorbox}

This schema is reconstructed from saved responses rather than from a
recovered literal judge prompt. The complete historical judge request was
not retained. In the representative GPT trace, proposer, selector, and
judge all use \texttt{gpt-5.2}, so judge acceptance may contain correlated
model errors and should not be interpreted as independently verified
physical correctness. No human-agreement estimate is available.

The maintained agent has a different acceptance rule: \texttt{Faithfulness=Yes}, confidence at least 0.70, ambiguity at most 0.30, and correct contact and orientation. Its evaluator can also return contact-angle and target-residual diagnostics, but the audited acceptance function does not use every such field as an additional hard gate. These current thresholds are not retroactively applied to the historical accepted counts.


\section{Additional Dataset and Diagnostic Analyses}
\label{app:additional_diagnostics}

This section provides complementary analyses of reference-tree
structure, procedural granularity, and release-threshold sensitivity.
These analyses clarify the scope and interpretation of the main
results without introducing additional model inference or
physical-simulation trials.


\subsection{Reference-Tree Structure and Procedural Units}
\label{app:tree_complexity}

\begin{figure*}[t]
    \centering
    \includegraphics[width=0.922\textwidth]
    {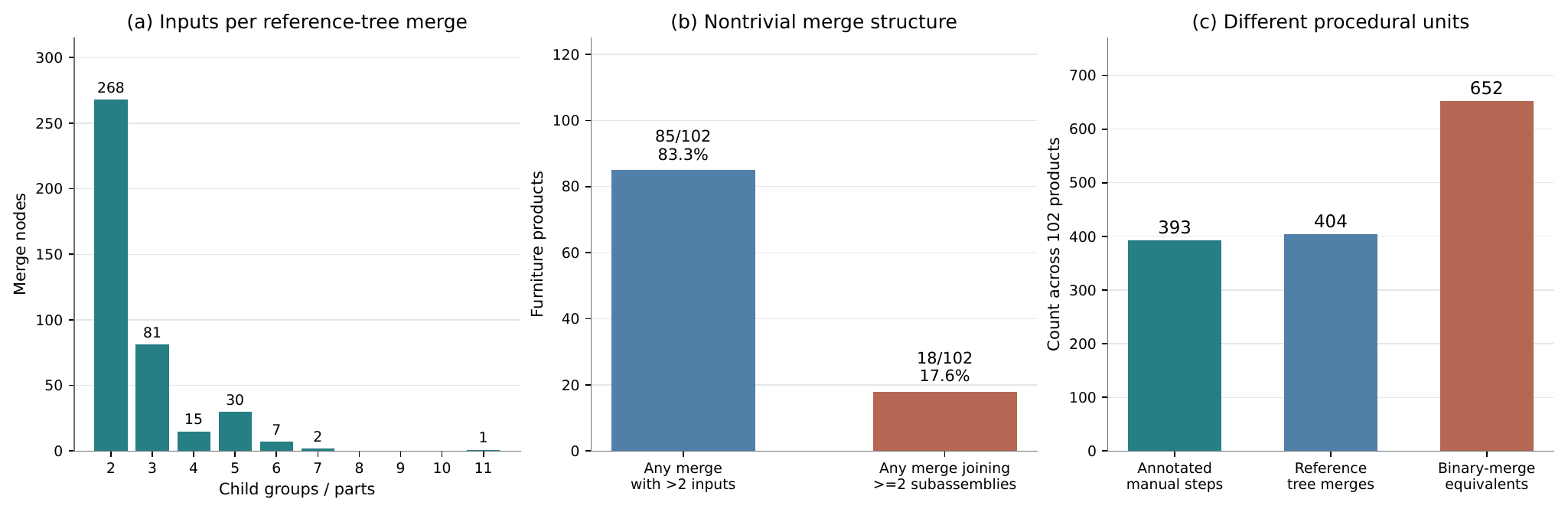}
    \caption{\textbf{Reference-tree structure and procedural granularity
    across 102 products.}
    Left: number of children associated with reference-tree merge nodes.
    Middle: products containing multiway merges or merges joining multiple
    subassemblies. Right: total annotated manual steps, reference-tree
    merges, and binary-merge equivalents.
    Binary-merge equivalents are combinatorial decompositions and should
    not be interpreted as verified robot actions or feasible trajectories.}
    \label{fig:app_tree_complexity}
\end{figure*}

Reference assembly trees do not in general define a unique sequence of
single-part executable actions. As shown in
Figure~\ref{fig:app_tree_complexity}, 85/102 products contain at least one
reference-tree merge with more than two inputs, and 18/102 contain a merge
that joins at least two non-leaf subassemblies. Across all products, the
metadata contain 393 annotated manual steps and 404 reference-tree merge
nodes, whereas decomposing multiway merges into binary equivalents yields
652 combinatorial merges. These counts demonstrate that manual steps,
hierarchical tree nodes, and atomic moving-part operations are distinct
procedural units.

\subsection{Sensitivity of the Release Diagnostic}
\label{app:release_sensitivity}

\begin{figure*}[t]
    \centering
    \includegraphics[width=0.9\textwidth]
    {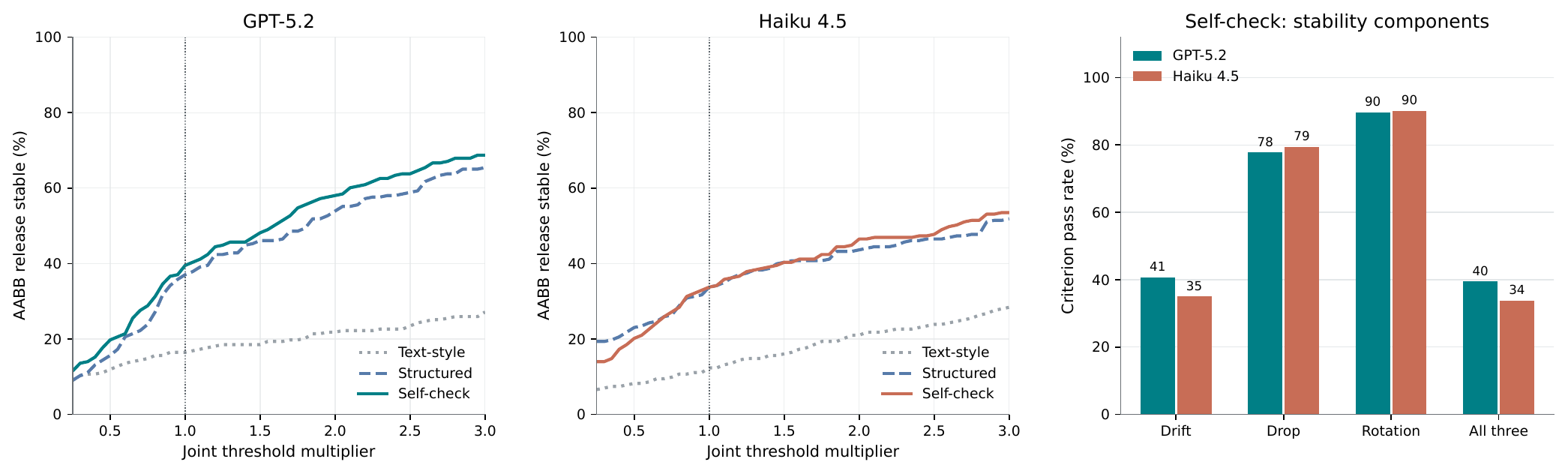}
    \caption{\textbf{Sensitivity of the AABB release diagnostic.}
    Left and center: release-stability rate when the three reported
    thresholds are jointly multiplied by a common factor; the vertical
    reference line marks the reported multiplier of $1.0$.
    Right: individual criterion pass rates for the self-check
    configurations. This is a post-hoc analysis of stored release records,
    not an additional physics rollout.}
    \label{fig:app_release_thresholds}
\end{figure*}

The box-based release label depends on explicit numerical thresholds and
should therefore not be interpreted as an invariant notion of physical
assembly success. Figure~\ref{fig:app_release_thresholds} jointly scales
the recorded drift, vertical-drop, and rotation thresholds around the
reported operating point. Stability rates vary continuously as these
thresholds are relaxed or tightened. The criterion-level breakdown further
shows that positional drift is the most restrictive of the three recorded
conditions for the self-check configurations, while rotation alone is
satisfied substantially more often.

This sensitivity does not invalidate the release diagnostic, but it
motivates treating it as a coarse support/stability probe rather than as
ground-truth connector-level success. In particular, the test does not
evaluate fastening, insertion constraints, mesh-level contact, gripper
interaction, or collision-free motion.

\subsection{IsaacLab AABB Release}
\label{app:release}
The physical record version is \textbf{isaaclab\_release\_aabb.v1} \cite{isaacsim}. For each selected moving-part prediction, each posed part is approximated as a world-axis-aligned cuboid. The fixed set becomes static collision geometry, and only the moving body is released dynamically. The simulation records starting/ending center positions, moving-box dimensions, and rotation drift.

Let $p_0,p_H$ denote the released box's initial and final centers, and $Q_0,Q_H$ its orientations. Define
\[
d=\|p_H-p_0\|_2,\qquad
z_{\mathrm{drop}}=\max(0,p_{0,z}-p_{H,z}),\qquad
\theta=\operatorname{GD}(Q_0,Q_H).
\]
The stable indicator is
\[
s=\mathbf1[d<0.05 \;\land\; z_{\mathrm{drop}}<0.05
\;\land\; \theta<12^\circ].
\]
All 1,701 records across the seven audited configurations satisfy this classification when recomputed from saved scalar fields. No new simulation is needed for this check.

\section{Prompt and Evaluation Protocol}
\label{app:prompt_protocol}

\setcounter{table}{0}
\renewcommand{\thetable}{A\arabic{table}}
\renewcommand{\theHtable}{A\arabic{table}}

This section documents the prompt schedule, model inputs,
intermediate outputs, and evaluation boundaries of the two
reasoning components: manual-to-structure recovery and
target-free pose refinement.

\subsection{Manual-to-Structure Prompt Schedule}
\label{app:manual_schedule}

The implementation uses a short sequence of calls whose outputs are passed forward as text state. The three colored blocks below follow the same stage colors used in the main pipeline figure.

\ProtocolBox{StageOne}{Stage 1 -- Part grounding (M1--M2)}{%
\textbf{M1 -- Part naming (\texttt{generate\_json}).}
The model first sees the numbered scene, available annotated scene views, and a manual preview page. It returns a JSON mapping from visible part IDs to part names.\\[3pt]
\textbf{M2 -- Semantic enrichment (\texttt{select\_material}).}
The M1 table is then shown together with the numbered scene/views and the full manual-page sequence. The output augments each part with a short functional explanation; this semantic table is the textual state passed to M3.
}

\ProtocolBox{StageTwo}{Stage 2 -- Anchor-guided planning (M3)}{%
\textbf{M3 -- Anchor planning (\texttt{planning\_anchor\_v2m}).}
One MLLM request receives the Stage-1 semantic table, the numbered scene, and the ordered numbered step masks. Within this single request, the prompt asks the model to identify boundary anchors, assign newly introduced parts to manual steps, track carried-over subassemblies, determine the assembly type, and write the corresponding textual plan. The saved response is passed to the tree-conversion stage.
}

\ProtocolBox{StageThree}{Stage 3 -- Tree conversion (M4)}{%
The Stage-2 text plan is converted to a nested integer-list representation of the assembly hierarchy. A rule-based parser is attempted first; when it does not produce a valid tree, the text-only fallback prompt \texttt{tree\_ikea\_manual} is used. No new images are introduced at this stage, and the resulting hierarchy is saved as \texttt{tree.json}.
}

\paragraph{Execution order.}
The same sequence in compact form is:
\CodeFileBox{Execution schedule}{latex/prompts/execution_order_print.txt}
The full available prompt snapshots are reproduced in the final prompt-snapshot section below.

\subsection{Pose-Refinement Prompt Schedule}
\label{app:pose_schedule}

The pose experiment is evaluated as a current-operation subtask: the current atomic operation and the preceding assembly state are treated as correct, while the MLLM is not shown the ground-truth target pose or a ground-truth target-state rendering. The preserved historical trace follows a proposer $\rightarrow$ render $\rightarrow$ selector $\rightarrow$ judge $\rightarrow$ feedback loop.

\paragraph{Initialization (P0).}
For each independently evaluated operation, the program constructs an operation record containing the moving part, fixed subassembly, manual-step context, and part semantics. This record remains fixed while the candidate pose is refined.

\ProtocolBox{PoseStage}{One refinement round (P1--P4)}{%
\begin{enumerate}
  \setlength{\itemsep}{2pt}
  \setlength{\parskip}{0pt}
  \item[\textbf{P1}] \textbf{Propose.} The proposer receives the current multi-view state, manual crop, operation record, and previous judge feedback when $r>0$, and predicts an incremental update $\delta T^r$.
  \item[\textbf{P2}] \textbf{Render.} The program applies $\delta T^r$ and renders the resulting after-state from the standard views. This transform/render step is programmatic rather than another MLLM call.
  \item[\textbf{P3}] \textbf{Select.} The visual selector compares the rendered candidate after-state(s) against the manual context and operation goal and selects the option to evaluate.
  \item[\textbf{P4}] \textbf{Diagnose.} The saved judge output records accept/reject, failure type, residual direction, and contact/orientation hints.
\end{enumerate}
}

\ProtocolBox{FeedbackStage}{Feedback rule (P5)}{%
If the round is rejected, the diagnosis from P4 is inserted into the next proposer request together with the newly rendered state. The loop then repeats from P1. The preserved experiment trace uses one candidate per round and at most three rounds.
}

\paragraph{Historical trace versus maintained source.}
The supplied handoff explicitly distinguishes this saved three-round experiment from the current refactored agent. The saved trace uses one proposer candidate per round, three refinement rounds, a VLM selector, and a separate stored judge diagnosis. The maintained source defaults to a different evaluator/proposer multi-path loop. We therefore use the saved prompts only to document the historical three-round behavior and do not substitute current default prompts for missing historical requests. The exact historical judge request template was not retained, so we report the preserved judge output rather than reconstructing missing prompt text.

\section{Information Boundary of the Two Subtasks}
\label{app:information_boundary}

The two evaluations intentionally expose different forms of state. The boundary is easier to read as a protocol statement than as an additional result table:

\ProtocolBox{NeutralStage}{Evaluation boundary}{%
\begin{itemize}
  \setlength{\itemsep}{3pt}
  \setlength{\parskip}{0pt}
  \item \textbf{Manual-to-structure.} The model receives the manual pages, numbered scene, and numbered step masks. The part semantic table is produced by the model and carried forward. The reference assembly tree is used only for evaluation.
  \item \textbf{Isolated pose evaluation.} The model receives the current manual crop, current rendered state, part semantics, and a correct current operation/state context. In the supplied pose adapter, this context is constructed from dataset tree/step metadata rather than by directly consuming the predicted \texttt{tree.json}.
  \item \textbf{Not exposed as targets.} The ground-truth target 6D pose and a ground-truth target-state rendering are not shown to the pose MLLM. Ground-truth pose is used only for scoring and, after scoring, to reset context for the next independently evaluated operation. Predicted after-state renders are the visual feedback used for subsequent correction rounds.
\end{itemize}
}

\section{Representative Manual-to-Structure Trace}
\label{app:manual_trace}

The supplied manual-side example contains four labeled structural parts. The retained \texttt{stage1\_output.json} is the enriched Stage-1 table after semantic grounding:

\begin{table}[H]
\centering
\small
\renewcommand{\arraystretch}{1.15}
\begin{tabularx}{\linewidth}{@{}c l >{\raggedright\arraybackslash}X@{}}
\toprule
\textbf{ID} & \textbf{Name} & \textbf{Functional explanation} \\
\midrule
0 & long side rail &
Forms the long side of the bed frame and supports the slatted bed base. \\

1 & slatted bed base &
Provides the flexible sleeping surface that supports the mattress. \\

2 & headboard/end frame &
Closes one end of the bed and provides the headboard structure. \\

3 & footboard/end frame &
Closes the opposite end of the bed and helps stabilize the frame. \\
\bottomrule
\end{tabularx}
\caption{Retained Stage-1 semantic table for the representative manual example.}
\label{tab:manual_stage1_example}
\end{table}

The next MLLM call receives this semantic table together with the numbered scene and ordered step masks. The structural portion of the saved Stage-2 output is:

\OutputFileBox{Retained Stage-2 structural output}{latex/prompts/manual_stage2_excerpt_print.txt}

The same response also contains natural-language assembly instructions. We omit those longer prose instructions here because the appendix is intended to expose the prompt/state flow rather than add another qualitative manual figure. The concrete retained example stops at the Stage-2 file; the supplied materials do not contain its final \texttt{tree.json}, so we do not fabricate a conversion result. The exact conversion prompt and its required nested-list output contract are provided in the full prompt-snapshot section below.

\section{Representative Three-Round Pose Trace}
\label{app:pose_trace}

We use the preserved Bench/applaro operation-0 trace because it contains all three rounds of actual proposer, selector, and judge records. The atomic operation moves part 2 (a side frame) relative to fixed part 1 (the seat panel), with the goal of attaching the frame upright to one end of the panel. This is a \emph{failure trace}, not a successful assembly example: all three rounds remain rejected, and the historical harness records ground-truth rescue only to construct later operation context.

\begin{table}[H]
\centering
\footnotesize
\setlength{\tabcolsep}{3pt}
\renewcommand{\arraystretch}{1.2}

\begin{tabularx}{\linewidth}{@{}
c
>{\raggedright\arraybackslash}p{0.22\linewidth}
>{\raggedright\arraybackslash}p{0.24\linewidth}
>{\raggedright\arraybackslash}X
@{}}
\toprule
\textbf{Round} &
\textbf{Feedback entering round} &
\textbf{Proposed incremental update} &
\textbf{Saved after-state diagnosis} \\
\midrule

0 &
None &
$\Delta t=[0,0,+1.50]$, identity rotation &
Rejected; \texttt{wrong\_side}; contact incorrect; orientation incorrect. Judge asks to move to a short end and rotate the frame upright. \\

1 &
Previous \texttt{wrong\_side} diagnosis and residual direction &
$\Delta t=[0,0,-0.45]$, $+90^\circ$ about world $Y$ &
Rejected; \texttt{wrong\_side} persists; frame remains separated from the panel end and not upright/perpendicular. \\

2 &
Updated \texttt{wrong\_side} diagnosis and contact/orientation hints &
$\Delta t=[0,0,+0.40]$, $+90^\circ$ about world $X$ &
Rejected; \texttt{wrong\_side}; the frame is still along the long side instead of seated flush on an end face. \\

\bottomrule
\end{tabularx}
\vspace{2pt}
\caption{Saved three-round refinement trace for Bench/applaro, operation 0. The feedback changes the next proposal, but does not resolve the contact topology within three rounds.}
\label{tab:pose_trace}
\end{table}

The key mechanism is that the previous diagnosis is inserted verbatim into the next proposer request. The retained round-1 request contains the following block:

\PromptFileBox{FeedbackStage}{Judge feedback inserted into the next proposer request}{latex/prompts/pose_feedback_excerpt_print.txt}

This trace illustrates why the paper reports both improvement and remaining failure modes: after-state feedback can redirect the update, yet a visually reasoned correction may continue to choose the wrong contact side or orientation.

\section{Inference Configuration for the Preserved Pose Trace}
\label{app:pose_config}

The archived Bench/applaro trace used for the appendix records the following client settings. These settings document this trace only; they should not be read as proof that every aggregate row labeled ``GPT-5'' in the main table used the identical API identifier or wrapper.

\begin{table}[H]
\centering
\small
\begin{tabular}{lccc}
\toprule
\textbf{Role} & \textbf{Model ID in trace} & \textbf{Max image side} & \textbf{Reasoning effort} \\
\midrule
Proposer & gpt-5.2 & 768 & low \\
Selector & gpt-5.2 & 512 & minimal \\
Judge & gpt-5.2 & 512 & minimal \\
\bottomrule
\end{tabular}

\vspace{5pt}

\caption{Model-client metadata stored with the representative historical trace. Proposer and selector/judge completion-token caps are 2048 and 512, respectively; image detail is recorded as high for all three roles.}
\label{tab:pose_trace_config}
\end{table}

The handoff package records one candidate per round and three rounds for this historical example. It also states that the complete historical judge prompt and system-message/API wrapper were not recovered. We therefore reproduce only prompt text and outputs that are actually preserved.

\section{Full Available Prompt Snapshots}
\label{app:prompt_snapshots}

\subsection{Part Naming Prompt (M1)}
\PromptFileBox{StageOne}{Prompt M1 -- Part naming}{latex/prompts/part_names_print.txt}

\subsection{Part Semantic-Enrichment Prompt (M2)}
\PromptFileBox{StageOne}{Prompt M2 -- Semantic enrichment}{latex/prompts/part_descriptions_print.txt}

\subsection{Anchor-Guided Planner Prompt (M3)}
\PromptFileBox{StageTwo}{Prompt M3 -- Anchor-guided planning}{latex/prompts/tree_v2m_print.txt}

\paragraph{Terminology note.} The historical M3 prompt uses the word ``ROOT for the final-step anchor part. In the paper and in the tree-conversion prompt, the graph-theoretic root denotes the completed assembly; we therefore refer to the former as the \emph{final-step anchor} in the main text.

\subsection{Text-to-Tree Conversion Prompt (M4)}
The following is reproduced as supplied, including its illustrative examples; it is a prompt snapshot rather than a validated ground-truth trace.
\PromptFileBox{StageThree}{Prompt M4 -- Text-to-tree conversion}{latex/prompts/tree_conversion_print.txt}

\subsection{Historical Initial Pose-Proposer Request}
This is the saved round-0 user prompt for Bench/applaro operation 0.
\PromptFileBox{PoseStage}{Pose proposer -- Initial request}{latex/prompts/pose_initial_print.txt}

\subsection{Historical Feedback-Conditioned Pose-Proposer Request}
This is the saved round-1 request after inserting the previous negative diagnosis.
\PromptFileBox{PoseStage}{Pose proposer -- Feedback-conditioned request}{latex/prompts/pose_feedback_print.txt}

\subsection{Historical After-State Selector Request}
\PromptFileBox{PoseStage}{After-state selector request}{latex/prompts/pose_selector_print.txt}

\section*{Limitations}

Although AssemState improves assembly-structure recovery and local 6D pose reasoning, several limitations remain. First, the manual-side evaluation relies on precomputed numbered step masks, while automatic mask generation from raw manuals is outside the scope of this work. Second, the closed-loop 6D pose refinement process is evaluated on independently assessed operations with correct atomic instructions and preceding assembly states. Therefore, the reported results do not measure error propagation across successive assembly steps or establish end-to-end assembly performance. Third, the refinement process relies on interactions between the VLM proposer and judge, which may accept visually plausible but geometrically incorrect poses. Errors involving collision, floating, mirror orientation, incomplete seating, and wrong-side attachment remain challenging. Finally, the simulation-based release tests use simplified collision geometry and provide only coarse physical stability diagnostics rather than verifying precise contact, fastening, or collision-free manipulation. These limitations highlight the need for more reliable spatial reasoning and physical verification.

\section*{Ethical Statement on LLM Assistance}

We primarily use GPT-5 as a tool for language refinement, including polishing text and improving clarity. All model-generated content is thoroughly reviewed and rewritten by human authors to ensure accuracy, originality, and adherence to research integrity standards.

\clearpage
\end{document}